\documentclass{article}
\usepackage[preprint]{neurips_2025}

\usepackage[utf8]{inputenc} % allow utf-8 input
\usepackage[T1]{fontenc}    % use 8-bit T1 fonts
\usepackage{hyperref}       % hyperlinks
\usepackage{url}            % simple URL typesetting
\usepackage{booktabs}       % professional-quality tables
\usepackage{amsfonts}       % blackboard math symbols
\usepackage{nicefrac}       % compact symbols for 1/2, etc.
\usepackage{microtype}      % microtypography
\usepackage{xcolor}         % colors
\usepackage{booktabs}
\usepackage{amssymb}
\usepackage{pifont}
\usepackage{multirow}
\usepackage{graphicx}
\usepackage{tabularx}

\title{Position: Intelligent Science Laboratory Requires the Integration of Cognitive and Embodied AI}

\author{%
  Sha Zhang \\
  University of Science and Technology of China\\
  \And
  Suorong Yang \\
  Nanjing University \\
  \AND
  Tong Xie \\
  University of New South Wales \\
  \And
  Xiangyuan Xue \\
  Shanghai Jiao Tong University \\
  \And
  Zixuan Hu \\
  Peking University \\
  \And
  Rui Li \\
  Harbin Institute of Technology \\
  \And
  Wenxi Qu \\
  Tongji University \\
  \And
  Zhenfei Yin \\
  The University of Sydney, Oxford \\
  \And
  Tianfan Fu \\
  Nanjing University \\
  \And
  Di Hu \\
  Renmin University of China \\
  \And
  Andres M Bran \\
  Swiss Federal Institute of Technology Lausanne  \\
  \And
  Nian Ran \\
  Shanghai Institute of Ceramics, Chinese Academy of Sciences \\
  \And
  Bram Hoex \\
  University of New South Wales \\
  \And
  Wangmeng Zuo \\
  Harbin Institute of Technology \\
  \And
  Philippe Schwaller \\
  Swiss Federal Institute of Technology Lausanne \\
  \And
  Wanli Ouyang \\
  Shanghai AI Laboratory \\
  \And
  LEI BAI \\
  Shanghai AI Laboratory \\
  \And
  Yanyong Zhang \\
  University of Science and Technology of China \\
  \And
  Lingyu Duan \\
  Peking University \\
  \And
  Shixiang Tang\thanks{Corresponding author} \\
  Shanghai AI Laboratory \& The Chinese University of Hong Kong\\
  \And
  Dongzhan Zhou$^{*}$ \\
  Shanghai AI Laboratory \\
}

\begin{document}

\maketitle

\begin{abstract}
Scientific discovery has long been constrained by human limitations in expertise, physical capability, and sleep cycles. The recent rise of AI scientists and automated laboratories has accelerated both the cognitive and operational aspects of research. However, key limitations persist: AI systems are often confined to virtual environments, while automated laboratories lack the flexibility and autonomy to adaptively test new hypotheses in the physical world.
Recent advances in embodied AI, such as generalist robot foundation models, diffusion-based action policies, fine-grained manipulation learning, and sim-to-real transfer, highlight the promise of integrating cognitive and embodied intelligence. This convergence opens the door to closed-loop systems that support iterative, autonomous experimentation and the possibility of serendipitous discovery.
In this position paper, we propose the paradigm of Intelligent Science Laboratories (ISLs)—a multi-layered, closed-loop framework that deeply integrates cognitive and embodied intelligence. ISLs unify foundation models for scientific reasoning, agent-based workflow orchestration, and embodied agents for robust physical experimentation. We argue that such systems are essential for overcoming the current limitations of scientific discovery and for realizing the full transformative potential of AI-driven science.
\end{abstract}
\section{Introduction}
\label{sec:intro}

Scientific discovery, the pursuit to expand the boundaries of human knowledge, has long been one of humanity’s most fundamental endeavors~\cite{gibbons1974roles}. Historically, scientific progress relied on researchers conducting experiments through manual labor and incremental intellectual effort, with knowledge painstakingly passed down via scholarly communication in language and figures. 
Recent advances in artificial intelligence and automation are transforming the scientific enterprise, moving it beyond manual experimentation toward a new era of AI scientists~\cite{huang2024position, lu2024ai, li2024academic, baek2024researchagent, da2024advancement, gao2024empowering} and automated laboratories~\cite{zhou2025multi, yue2025ir, zhao2023robotic, zhu2022all}. 
Specifically, the AI scientist paradigm typically works as the ``cognitive layer'' of science on the basis of foundation models and knowledge reasoning, while the autonomous science laboratories play a role as the ``action layer'' by automating the complex experiment operations. 

Despite great recent success, the current AI Scientist paradigm and the autonomous laboratory are still not sufficient to fully automate scientific research. On the one hand, although current AI Scientist paradigms~\cite{gottweis2025towards, microsoft2025discovery} can accelerate the “cognitive layer” of science by rapidly analyzing extensive datasets~\cite{cox2023prediction, howarth2020dp4, rankovic2025gollum}, generating hypotheses~\cite{wang2023hypothesis, baek2024researchagent}, analyzing experimental results~\cite{shields2021bayesian,hardwick2020digitising}, writing academic papers~\cite{elbanna2024exploring,moosavi2021scigen,imran2023analyzing} and even autonomously proposing new research directions~\cite{zenil2023future,sparkes2010towards}, they remain largely confined to data- and model-driven \emph{virtual} environments, lacking the capacity for real-world physical experimentation and the ability to translate cognitive insights into physical action. 
On the other hand, automated laboratory initiatives~\cite{tom2024self}—such as the ``Xiaolai'' lab~\cite{zhu2022all}—have significantly advanced the ``action layer'', which remarkably improves the efficiency, accuracy, and reproducibility of laboratory operations. Yet, these systems typically rely on rigid, pre-defined workflows in \emph{physical} environments and offer limited generalization or autonomous decision-making, making them inadequate for addressing the complexity and dynamism of modern scientific inquiry. 
Although recent advances in embodied~\cite{chi2023dp, zhao2023act} have significantly advanced the state of general robotic manipulation and adaptive control, current research predominantly targets standard benchmarks involving rigid, opaque objects and relatively straightforward interaction scenarios. However, laboratory environments present a distinct set of challenges, such as the precise manipulation of transparent materials, fine-grained liquid handling, and robust operation under intricate physical and safety constraints. Existing embodied AI systems lack the specialized capabilities required to address these domain-specific demands. Bridging this substantial gap necessitates a new paradigm that integrates advanced cognitive reasoning with embodied agents specifically designed to operate reliably within the complex and nuanced settings characteristic of real-world scientific laboratories.

Building upon this insight, we propose \textbf{intelligent science laboratory}, a multi-layered collaborative architecture. Our system unifies the powerful reasoning and planning capabilities of AI scientists with the robust physical execution of automated laboratories, thus enabling end-to-end, closed-loop intelligent science laboratories and accelerating scientific discovery. By tightly coupling the Foundation Model, Agent Layer, and Embodied Layer, our approach supports hypothesis generation, experimental design, physical operation, feedback-driven adaptation, and continual self-improvement. This architecture allows for efficient reasoning and workflow planning in virtual space, as well as adaptive, dynamic, and lifelong learning in real-world environments, positioning it as a core engine for the next paradigm shift in scientific research.

In short, this paper posits that \textbf{Intelligent Science Laboratories (ISLs) are essential for transcending the current limits of scientific discovery, by deeply integrating cognitive and embodied intelligence} within end-to-end, closed-loop frameworks to fully realize the transformative potential of AI-driven science.

The paper is structured as follows: Section~\ref{sec:what} provides precise definitions and scope for our position. Section~\ref{sec:why} elaborates on the limitations of existing approaches and motivates the need for our proposed framework. We then systematically present the three-layer architecture and its key enabling technologies in Section~\ref{sec:framework}, followed by case studies that illustrate its potential for addressing complex scientific tasks. Section~\ref{sec:challenges} discusses outstanding challenges and open questions. Finally, Section~\ref{sec:con} offers concluding remarks, and Section~\ref{sec:impact} reflects on the broader impact of this research.

\section{What is Intelligent Science Laboratory}
\label{sec:what}

The Intelligent Science Laboratory (ISL) is a new research paradigm that aspires to fully automate the process of scientific discovery~\cite{butakova2021data}. Envisioned as an autonomous, robust, and self-improving platform, an ISL is capable of generating hypotheses, designing and executing experiments, analyzing data, and adapting its strategies, all without or with limited human intervention. 

At its core, the ISL unifies three key components within a closed-loop architecture: foundation models, agentic reasoning, and embodied automation.
These layers contribute distinct capabilities that, when combined, enable continuous optimization from hypothesis generation to experimental validation.
First, foundation models act as the cognitive core~\cite{taylor2022galactica,bran2023chemcrow,pan2024unifying}, capable of understanding and reasoning over heterogeneous scientific inputs, ranging from text and spectra to reaction graphs and simulation outputs~\cite{taylor2022galactica,bran2023chemcrow,wang2019smiles,gupta2022matscibert,beltagy2019scibert}.
They enable fast adaptation to in-depth, specific scientific tasks through closed-loop learning, while retaining their original capabilities, such as handling multidisciplinary knowledge, planning, reasoning, etc.
Second, the agent layer functions as a strategic orchestrator. It decomposes high-level goals into modular subtasks, dynamically allocates resources (e.g., instruments, models, reagents), and employs adaptive decision-making frameworks, such as reinforcement learning and Bayesian optimization~\cite{gelbart2014bayesian,settaluri2021automated}.
This layer ensures that the ISL continuously adapts to experimental outcomes and reprioritizes directions with the highest expected utility.
Third, the embodied automation layer closes the loop by physically executing experimental plans in either simulated or real environments. Through the Real2Sim2Real cycle~\cite{delgado2023research,coley2020autonomous,hu2023toward,yager2023autonomous}, the system first validates hypotheses virtually, then transitions to wet-lab execution with real-time sensing, safety monitoring, and error recovery. This dramatically reduces operational risk and resource waste.
The long-term vision for ISLs is to reach ---and ultimately surpass--- the comprehensive experimental capabilities of human scientists: autonomously identifying novel research directions, rapidly iterating on experimental protocols, and adapting to unforeseen challenges across diverse scientific domains.

To capture the progressive autonomy and intelligence of an ISL, we introduce a four-level taxonomy as shown in Table~\ref{tab:lab-levels}: from basic script-driven automation (Level 0), through context-aware local intelligence (Level 1) and closed-loop autonomous experimentation (Level 2), up to continual learning and self-evolution (Level 3). Each level marks a qualitative leap in autonomy, adaptability, and scientific capability.

\begin{table}[t]
\centering
\small
\caption{Levels of autonomy and intelligence in the AI-Powered Intelligent Lab framework, detailing the progression in Foundation Model, Agent, Embodied Layer, and Overall System Capability.}
\renewcommand{\arraystretch}{1.3}
\begin{tabular}{m{0.6cm} | m{2.1cm} | m{2.5cm} | m{3.2cm} | m{3.2cm}}
\hline
\textbf{Level} & \textbf{Foundation Model Layer} & \textbf{Agent Layer} & \textbf{Embodied Layer} & \textbf{Overall System Capability} \\
\hline
0 
& None or narrow, task-specific models.
& None.
& Executes predefined scripts via basic hardware; limited perception, lacks planning or adaptation.
& Deterministic, manually programmed, and confined to static environments. \\
\hline
1 
& Multimodal perception with basic scientific understanding; limited domain reasoning.
& Agent with local planning, simple adaptation, and collaborative abilities.
& Robotic platforms capable of local planning, short-range composite tasks, and basic sensor-driven adaptation.
& Executes medium-complexity tasks with limited generalization; requires human support for novel conditions. \\ 
\hline
2 
& Deep scientific knowledge; reasoning, analysis, and decision-making.
& Intelligent agent with scheduling, real-time feedback integration, and dynamic adaptation.
& Robotic systems execute multi-step protocols with integrated perception, navigation, and manipulation, guided by real-time feedback for adjustment, recovery, and safety.
& End-to-end closed-loop automation across sim and real labs; reaches researcher-level proficiency in known domains with minimal human oversight.\\
\hline
3
& Continually learning, self-adaptive, and innovative models.
& Self-evolving agent with complex planning, meta-learning, and autonomous optimization.
& Adaptive systems in dynamic environments support continual learning, calibration, tool switching, and generalization to novel tasks.
& Fully autonomous, self-improving discovery loop; surpasses human experts in specific domains. \\
\hline
\end{tabular}

\label{tab:lab-levels}
\end{table}

\section{Why Do We Need the Intelligent Science Laboratory}
\label{sec:why}

Traditional scientific labs rely heavily on human intuition and manual experimentation.
Hypotheses are typically generated based on expert heuristics and domain knowledge, which, while effective in well-understood domains, are difficult to scale across different areas.
The subsequent translation of hypotheses into experimental protocols is labor-intensive, often requiring fine-grained, domain-specific expertise~\cite{hickman2023self}.
Execution cycles are slow and costly, taking days or months, while failed experiments rarely feed back into the system in a structured way~\cite{coutant2019closed,macleod2020self,flores2020materials}. 
While the rise of high-throughput and robotic platforms has improved raw throughput, existing automation remains largely pre-scripted, brittle, and task-specific~\cite{eyke2021toward,coley2019robotic,abolhasani2023rise,coley2020autonomous}. These systems lack the adaptive intelligence to modify plans on-the-fly, reason over noisy outputs, or autonomously explore novel problem settings. Furthermore, modern scientific problems increasingly involve multi-objective trade-offs, noisy or sparse feedback, and complex design spaces that cannot be efficiently navigated through human intuition or trial-and-error alone.

The ongoing transition to ISLs has begun to reshape scientific workflows in many disciplines~\cite{hickman2023self,abolhasani2023rise,stach2021autonomous,saikin2019closed,pollice2021data,thomas2021applications}, offering early evidence of their potential despite current systems being far from fully autonomous~\cite{coley2020autonomous,coley2019robotic,hickman2023self,delgado2023research}.
Crucially, these three layers interact recursively. For instance, experimental results are immediately interpreted by the foundation model to refine its internal representations, while the meta-agent updates its action policies and workflow plans accordingly. This tight integration allows ISLs to accelerate discovery, reduce experimental burden, and scale across domains where traditional automation remains inflexible or inefficient.

\section{Conceptual Framework}
\label{sec:framework}

\begin{figure}[t]
\centering
\includegraphics[width=\linewidth]{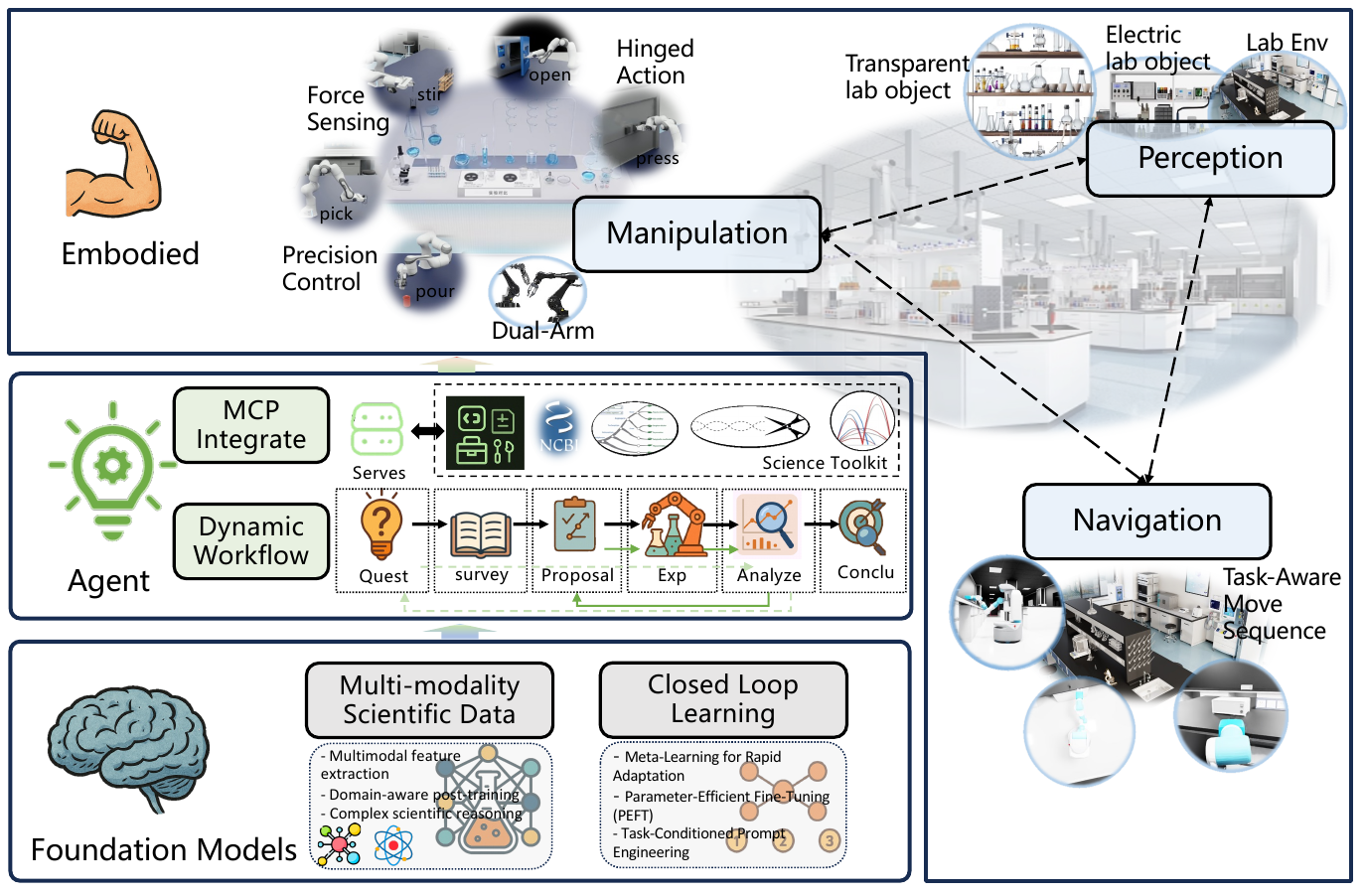}
\caption{
Overview of the \textbf{Intelligent Science Lab} framework. 
The framework consists of three collaborative layers: (1) Foundation Models provide multi-modal scientific knowledge representation and closed-loop learning capabilities, supporting complex reasoning and domain adaptation; (2) Agent Layer dynamically orchestrates scientific workflows—including hypothesis generation, literature review, experimental planning, execution, and analysis—while integrating model/toolkit via MCP integration; (3) Embodied Layer realizes robust physical interaction through advanced perception, navigation, and manipulation modules, enabling precise, adaptive operations in real-world laboratory environments. Arrows indicate information flow and coordination among perception, navigation, and manipulation to support fully autonomous, end-to-end scientific discovery.
}
\label{fig:framework}
\vspace{-3mm}
\end{figure}

We present a formal framework for the Intelligent Science Lab, in which three interdependent layers, namely Foundation Model Layer, Agent Layer, and Embodied Layer, collectively realize the standard scientific workflow in a closed-loop, end-to-end manner. Figure~\ref{fig:framework} illustrates the overall architecture and its alignment with canonical research stages.

\subsection{Architectural Overview}
The proposed framework interleaves three ISL layers to cover the full spectrum of procedures involved in scientific research. We abstract the scientific discovery process into six core capabilities: (1) problem formulation, (2) literature retrieval and analysis, (3) solution proposal and experimental design, (4) experimental execution, (5) data analysis, and (6) conclusion generation. 
Each layer contributes specific competencies and progressively increases system autonomy across four maturity levels (Level 0–3):
\begin{itemize}
    \item \textbf{Foundation Model Layer} acts as the laboratory’s cognitive core, responsible for hypothesis generation, literature synthesis, and data-driven modeling.
    \item \textbf{Agent Layer} functions as the strategic planner, converting hypotheses into executable workflows, scheduling resources via a unified Model-Calling Protocol, and managing closed-loop feedback.
    \item \textbf{Embodied Layer} comprises the physical (or simulated) execution layer, providing perception, navigation, and manipulation capabilities to carry out and monitor experiments under a Real-to-Sim-to-Real paradigm.
\end{itemize}

These three ISL layers co-evolve in a synergistic manner, progressively enhancing the intelligence and autonomy of the scientific research laboratory. 

\subsection{Foundation Model Layer}
To enable generalizable and adaptive intelligence in AI-powered scientific discovery, our framework is grounded in two core capabilities of foundation models: (1) understanding multi-modality scientific data, and (2) supporting closed-loop learning for rapid adaptation.

\textbf{Multi-modality Scientific Data.} Scientific environments inherently involve diverse data types—including structured measurements, images, natural language protocols, and simulation outputs. Our system leverages foundation models pre-trained on large-scale multimodal corpora, followed by domain-aware post-training to align with specific scientific tasks. These models perform multimodal feature extraction to unify heterogeneous inputs and enable complex scientific reasoning, such as inferring causal relationships across modalities or integrating experimental protocols with observed results.

\textbf{Closed-Loop Learning.} Scientific discovery is often constrained by limited data availability, high experimental cost, and evolving task demands, making few-shot learning essential. To address these challenges, we adopt a closed-loop learning paradigm, which continuously refines models through cycles of hypothesis generation, experimental feedback, and model updating. This looped structure enables the system to learn efficiently from sparse feedback and adapt to new tasks with minimal supervision.
Specifically, we integrate (i) meta-learning strategies for rapid task adaptation~\cite{finn2017model,andrychowicz2016learning}, (ii) parameter-efficient fine-tuning (PEFT) methods for low-cost updates~\cite{hu2022lora,houlsby2019parameter}, and (iii) task-conditioned prompt engineering that dynamically tailors model behavior to different experimental goals. Together, these techniques empower the system to operate in real-world scientific settings, where closed feedback loops and few-shot adaptation are critical for practical deployment.

\subsection{Agent Layer}
Scientific discovery often requires the coordination of heterogeneous models, tools, and agents across tasks such as hypothesis generation, experimental design, and data interpretation. However, current multi-agent scientific systems typically rely on predefined workflows, which lack the flexibility to generalize across domains with varying collaboration patterns. To address this limitation, we introduce a Meta-Agent responsible for automatic workflow generation and adaptation.

\textbf{Automatic Workflow Design.} Considering the diversity and flexibility of workflows across scientific tasks, we propose that the agent layer should not only execute predefined workflows but also be capable of autonomously designing them. We denote this capability as the meta-agent. The meta agent observes the task context, such as problem type, available modalities, and domain-specific constraints, and dynamically composes a chain of specialized agents to form an executable workflow. Instead of relying on fixed templates, it leverages a library of modular capabilities and task schemas to assemble workflows on the fly. This enables the system to generalize to novel scientific problems and reconfigure itself in response to intermediate results or unexpected experimental feedback.

\textbf{MCP-like Integration of Proprietary Models.} Modern scientific advances often hinge on the use of high-performance, domain-specific models such as AlphaFold~\cite{abramson2024accurate}. To make such models interoperable within a multi-agent system, we propose a Model-Call Protocol abstraction layer. This layer provides a unified interface for invoking proprietary or non-open-source models, treating them as callable functions with defined I/O signatures and side-effect semantics. By wrapping these models as agents with callable APIs, the system can seamlessly orchestrate both open-source foundation models and domain-specific expert models, reducing the need for building solutions from scratch.

\subsection{Embodied Layer}
To execute scientific procedures in real-world laboratories, the AI-powered system must possess robust embodied intelligence that tightly couples perception, navigation, and manipulation. We introduce an Embodied Layer composed of three core modules—Perception, Navigation, and Manipulation—designed to support complex and sensitive experimental actions in physical environments.

\textbf{Perception}: Laboratory settings present unique challenges for perception, including recognizing transparent or reflective labware, detecting fine-grained liquid levels, and interpreting complex spatial arrangements under domain-specific lighting and occlusion. Our perception module incorporates vision-language models and depth sensing to identify lab-specific objects and understand their functional roles in scientific workflows.

\textbf{Navigation}: Precise and task-aware mobility is crucial for orchestrating multi-step experimental procedures, particularly in constrained bench-top environments. The navigation module enables task-aware move sequencing, leveraging spatial reasoning and motion planning to move between stations, avoid obstacles, and position actuators with sub-centimeter accuracy.

\textbf{Manipulation}: Scientific manipulation requires high precision and sensitivity, particularly in tasks such as liquid handling, pipetting, or force-sensitive placement of samples. The manipulation module supports both discrete and continuous control strategies, integrating tactile feedback and visual servoing to ensure safe and accurate task execution.

\textbf{Integrated Perception-Action Loop}: The Embodied Layer is designed as a closed-loop system, where perception continuously informs both navigation and manipulation. For example, accurate detection of a reagent’s liquid level (perception) guides pipette insertion depth (manipulation), while recognizing spatial layout constraints informs path planning (navigation). This tri-modular synergy enables embodied agents to perform real-world wet-lab operations with reliability and autonomy.

\subsection{System Levels and Layer Mapping}
To systematically assess the progression of autonomy and intelligence in the Intelligent Science Lab, we define a four-level taxonomy, ranging from basic automation (Level 0) to fully self-evolving closed-loop discovery (Level 3). Each level is characterized by the integrated capabilities of the three principal layers: Foundation Model, Agent, and Embodied Layer.

\textbf{Level 0: Automation Lab.}
The system executes predefined scripts using basic automation hardware, without perception, reasoning, or adaptation. All tasks are performed deterministically within a closed environment, requiring continuous human oversight. The Foundation Model and Agent layers are either absent or limited to narrow, task-specific functions. The Embodied Layer operates deterministically in static environments, with all decision-making and oversight performed by humans.

\textbf{Level 1: Context-Aware Local Intelligence.}
The Foundation Model incorporates basic multimodal perception and limited domain reasoning. The Agent layer supports local planning and simple adaptation, while the Embodied Layer is capable of medium-complexity workflows and short-range context adaptation. Generalization and memory remain weak, necessitating frequent human intervention for novelty and environmental changes.

\textbf{Level 2: Closed-Loop Scientific Experimentation.}
The laboratory operates as a fully automated, end-to-end closed-loop system, where intelligent agents dynamically orchestrate multi-step experimental workflows in response to real-time feedback. Enabled by a domain-adapted foundation model, the system demonstrates researcher-level proficiency in familiar scientific domains, with human involvement primarily limited to supervisory oversight. At the core, the Foundation Model supports deep scientific reasoning and facilitates continual learning through closed-loop adaptation. The Agent Layer coordinates complex workflows by integrating real-time data streams and adjusting experimental plans on the fly. The Embodied Layer—comprising integrated modules for perception, navigation, and manipulation—autonomously executes and refines intricate experimental protocols across both simulated and physical environments. Together, these components form a tightly coupled architecture that minimizes human intervention while maintaining scientific rigor and adaptability.

\textbf{Level 3: Continual Learning and Self-Evolution Scientific Intelligence Laboratory.}
The system demonstrates robust and sustained autonomy through tightly integrated closed-loop feedback mechanisms, enabling continuous learning from experimental outcomes and dynamic adaptation across tasks and environments. In well-characterized scientific domains, it has the potential to surpass human experts in efficiency, throughput, and operational safety. All three architectural layers exhibit domain-specific self-improving capabilities: the Foundation Model continually ingests and models experimental data, supporting cross-modal representation learning, symbolic reasoning, and generalization to novel tasks; the Agent Layer performs multi-step planning, task decomposition, and dynamic scheduling, leveraging meta-learning to optimize policies in response to evolving goals and environmental conditions; and the Embodied Layer, equipped with modules for perception, navigation, and manipulation, autonomously executes complex protocols with real-time calibration, tool switching, and fault recovery in partially observable and dynamic settings. Together, these components form a unified, self-adaptive architecture that enables fully autonomous, end-to-end scientific discovery, advancing a new paradigm in experimental research.

\subsection{Application}
ISLs are rapidly expanding their impact across scientific domains.
In drug discovery, ISLs enable closed-loop optimization across key stages such as target identification, compound generation, and assay refinement~\cite{mullard2022r,  jayatunga2022ai,  stokes2020deep}. 
By integrating generative deep learning~\cite{zhavoronkov2019deep, ingraham2023illuminating, watson2023novo}, protein structure prediction~\cite{jumper2021highly,ren2023alphafold}, and automated synthesis platforms~\cite{schneider2018automating}, ISLs can iteratively propose, synthesize, and evaluate candidate molecules with minimal human intervention. Representative successes include GNN-driven antibiotic discovery~\cite{stokes2020deep} and AlphaFold-guided design for dark proteins~\cite{ren2023alphafold}. Despite these advances, ISLs in drug discovery face challenges related to multi-scale biological complexity and noisy experimental feedback~\cite{saikin2019closed}. Nonetheless, stage-specific deployments—particularly in hit-to-lead optimization and compound profiling—are accelerating the shift toward autonomous pharmaceutical pipelines~\cite{ jayatunga2022ai,gorgulla2020open}.
In the domain of structural materials, while fully autonomous ISLs remain rare~\cite{miracle2017new}, systems like HT-READ demonstrate the integration of automated alloy synthesis with high-throughput characterization~\cite{vecchio2021high}. DeCost et al. employed an autonomous electrodeposition platform combined with active learning to discover corrosion-resistant Al-Ni-Ti alloys~\cite{decost2022towards}, while Gongora et al. used simulation-guided Bayesian optimization to design 3D-printed structural materials~\cite{gongora2020bayesian}.
Energy storage research has seen rapid ISL adoption for material discovery and process optimization. Electrolab enables automated formulation and electrochemical screening of redox flow battery electrolytes~\cite{oh2023electrolab}, and Hickman et al. developed a low-cost platform coupling synthesis with electrochemical analysis for redox-active complexes~\cite{hickman2025atlas}. In lithium-ion batteries, Bayesian optimization has accelerated electrolyte discovery, while the CASH platform automates solid-state synthesis and resistance testing~\cite{dave2020autonomous, svensson2023robotised,shimizu2020autonomous, fatehi2023critical}. 
In electrocatalysis, Fatehi et al. implemented an ISL to identify earth-abundant oxygen evolution catalysts through proxy stability tests in a closed-loop design framework~\cite{wasmus1999methanol}. These platforms exemplify ISLs' ability to navigate high-dimensional, multi-objective landscapes in energy research.
Beyond domain-specific applications, ISLs have also advanced chemical synthesis and catalysis. Platforms such as IBM RoboRXN~\cite{granda2018controlling,o2021ai} and MIT’s automated organic synthesis systems integrate predictive models with robotic execution to perform complex multi-step reactions~\cite{wang2020automated,ekins2024lab}. 
In inorganic and coordination chemistry, ISLs have autonomously explored new metal-ligand complexes and surface-active species by combining automated liquid handling with spectroscopy and electrochemical analysis~\cite{zhao2025artificial, soldatov2021self,porwol2020autonomous, kowalski2023automated, laws2024autonomous, pabloaffordable}. 
Furthermore, ISLs serve as testbeds for methodological innovation, enabling systematic evaluation of optimization strategies~\cite{laws2024autonomous,pablo2025affordable,kowalski2023automated}, such as Bayesian optimization, reinforcement learning, and meta-learning, for accelerating experimental convergence~\cite{liang2021benchmarking, hickman2025atlas, feurer2018scalable}.

\section{Challenges}
\label{sec:challenges}
\subsection{Foundation Model Layer Challenges}

\textbf{Native Multimodal Scientific Data Understanding.} A fundamental challenge for autonomous scientific laboratories lies in the native representation and interpretation of complex, domain-specific data modalities produced by modern experimentation, including but not limited to spectra, chromatograms, mass spectra, and biomolecular sequences. These data are typically high-dimensional, heterogeneous, and characterized by significant noise and modality-specific artifacts~\cite{yang2024clip,maffettone2023missing,dou2023machine}. Current foundation models and general-purpose multimodal architectures (e.g., Perceiver~\cite{jaegle2021perceiver}, Multimodal Transformers~\cite{li2024multimodal}) are not inherently equipped to capture the nuanced physical, chemical, and biological semantics embedded within such scientific signals~\cite{zhang2025scientific}.

To enable robust end-to-end automation, it is imperative to develop foundation models capable of directly encoding and reasoning over these specialized modalities. This entails significant methodological challenges:
(i) Designing unified and expressive embedding spaces that faithfully preserve modality-specific structure and information content;
(ii) Establishing robust cross-modal alignment mechanisms to coherently integrate diverse instrument outputs into a shared latent space; and
(iii) Ensuring semantic integrity and interpretability such that downstream scientific reasoning and hypothesis generation remain physically and chemically meaningful.

\textbf{Scientific Reasoning and Executable Protocol Planning.}
While large models have demonstrated impressive capabilities in mathematical reasoning and code generation, translating high-level scientific hypotheses into machine-actionable experimental protocols remains an open challenge. Foundation models must propose not only novel research ideas but also detailed, step-by-step procedures parameterized for robotic execution. Moreover, the reasoning process should also incorporate effective analysis of feedbacks from the simulator or real-world experiment data. Inspired by cognitive Slow System/ Fast System fusion, we advocate a dual-pathway design: a lightweight “intuition” module rapidly filters candidate hypotheses, followed by a rigorous reasoning module that refines parameter values and procedural steps into fully specified, executable protocols. This approach ensures both creativity and feasibility, bridging the gap between theoretical insight and practical laboratory implementation.

\subsection{Agent Layer Challenges}

\textbf{Efficient Agent Specialization.} While foundation models demonstrate impressive general-purpose reasoning capabilities, they often fall short when tackling fine-grained or domain-specific scientific tasks. Relying on foundation models alone to solve all subtasks can be computationally expensive and inefficient, especially when task complexity varies significantly. To this end, efficient agent specialization becomes crucial. This involves enabling agents to rapidly adapt to specific tasks, ideally through self-improving mechanisms that require minimal human supervision. While frameworks such as DSPy~\cite{khattab2023dspy} and GRPO~\cite{shao2024deepseekmathpushinglimitsmathematical} advance toward this goal, current techniques still struggle to provide the level of specialization and scalability required for real-world scientific workflows.

\textbf{Sophisticated Tool Usage.} AI agents rely on tool invocation through well-defined APIs, where tools perform narrowly-scoped operations based on structured inputs. However, in scientific domains, tool usage often goes far beyond simple API calls. Scientific tools can include database systems requiring complex query generation, specialized software with steep learning curves, interactive graphical user interfaces, and even physical instrumentation. This diversity demands advanced capabilities from agents, including domain expertise, adaptive planning, and the ability to engage in multi-step or closed-loop interactions. Current agent architectures are generally ill-equipped to handle such sophisticated toolchains, and significant innovations are needed to bridge this gap.

\textbf{Effective Multi-agent Collaboration.} Scientific inquiry is inherently multi-faceted, involving tasks such as data collection, hypothesis formulation, experimental design, and result interpretation—many of which can and should be executed in parallel by specialized agents. Realizing this decentralized workflow requires seamless multi-agent collaboration, where agents must not only coordinate responsibilities but also exchange information accurately and meaningfully. However, existing multi-agent systems often suffer from inefficient communication protocols, lack of contextual awareness, and failure to maintain shared understanding. Given the high stakes and low error tolerance of scientific research, these shortcomings pose serious challenges. Designing robust collaboration mechanisms tailored to research-grade multi-agent systems remains a pressing and open research problem.

\subsection{Embodied Layer Challenges}

\textbf{Real-world Robot. }
% 真机的challenge
In the field of Embodied AI, training and deploying models using real-world robot data faces numerous challenges. Firstly, the process of collecting real-world data is highly costly and inefficient. Unlike simulation environments, obtaining diverse and extensive data from physical scenarios requires significant expenditures on hardware, experimental spaces, and human resources~\cite{lim2024open,bu2025agibot}. Moreover, training is constrained by real-time interactions in physical environments, making large-scale parallelization difficult. Secondly, robots face inherent limitations related to hardware and sensor performance. High-quality multimodal sensors are expensive, and inconsistent data standards across different sensor brands and types complicate multimodal integration. Additionally, practical deployments demand real-time data processing, posing further computational challenges. Thirdly, the complexity and unpredictability of real-world environments greatly elevate system difficulty. Numerous physical phenomena, such as friction, resistance, and lighting variations, cannot be accurately modeled, and dynamic factors like pedestrians and obstacles require robots to possess substantial adaptability and robustness. Furthermore, hardware diversity and inconsistent control mechanisms across robot platforms pose significant challenges to data standardization and reproducibility, hindering model generalization across different systems. Lastly, Embodied AI systems deployed in real-world environments also face security threats, particularly adversarial attacks, where subtle and malicious input perturbations can mislead robot perception and decision-making, posing severe safety risks.

\textbf{Real‐to‐Sim‐to‐Real Gap. }
High‐fidelity digital twins can dramatically reduce risk and cost by validating protocols in silico before wet‐lab execution. However, when the discrepancy between simulation and reality—commonly known as the Sim‐to‐Real gap—exceeds tolerable bounds, protocols that succeed in simulation often fail on physical hardware. Quantifying and bounding this gap across diverse experimental modalities (fluidics, optics, thermal control) remains an open problem. Without systematic metrics and calibration strategies, autonomous labs will struggle to generalize simulation‐optimized workflows to real‐world conditions.

\textbf{Heterogeneous Sensor Fusion and Global State Estimation. }
Modern automated platforms integrate a multitude of sensors—spectrometers, mass analyzers, force/torque probes, high‐resolution cameras, and thermal imagers—each producing data streams with different formats, sampling rates, and time stamps. Fusing these heterogeneous inputs into a coherent, real‐time estimate of the experiment’s global state is nontrivial. Current middleware solutions lack standardized interfaces and latency guarantees for high‐dimensional sensor fusion, leading to inconsistent state representations that undermine closed‐loop control.

\section{Conclusion and Call to Action}
\label{sec:con}
In this paper, we introduce Intelligent Science Laboratories (ISLs) as a unified paradigm that integrates foundation models, agentic reasoning, and embodied automation into an end-to-end, closed-loop architecture.
By bridging the gap between cognitive and physical intelligence, ISLs lay the groundwork for a new era of scalable, adaptive, and self-improving scientific discovery.
While this vision is compelling, the journey towards realizing ISLs at scale is fraught with challenges: scientific understanding, protocol-level reasoning and execution, agent coordination, and robust real-world embodiment.
Yet, these challenges also open up \textit{vast opportunities} for innovation across AI, embodied systems, and scientific methodology.
We hope this work offers a conceptual roadmap for ongoing and future research in ISLs, positioning ISLs as both a blueprint and a call to action for the broader research community driving the next generation of AI-driven science. 

\section{Impact Statement and Alternative Views}
\label{sec:impact}

\textbf{Equity and Fairness. }
The high capital and technical requirements of ISLs risk widening the gap between well-funded institutions and resource-limited labs. Without careful subsidy, open‐source tooling, or shared infrastructure models, under-resourced groups may be shut out of cutting-edge experimental capabilities.

\textbf{Domain Bias and Research Diversity. }
ISLs often focus on domains with abundant data and tractable protocols (e.g., standard chemical syntheses), potentially neglecting high-impact but data-sparse areas such as superconductivity or novel materials where automation struggles. This could channel scientific investment into ‘low-hanging fruit’ at the expense of breakthrough discoveries.

\textbf{Workforce Transformation and Skill Development. }
Widespread ISL adoption will reshape the experimental workforce: routine tasks shift to machines, while demand grows for lab engineers, data scientists, and AI specialists. This transition requires new training programs, interdisciplinary curricula, and lifelong learning paths to equip scientists with the skills to design, audit, and interpret AI-driven experiments.

\textbf{Data Governance, Transparency and Reproducibility. }
Automated labs generate vast, high-velocity datasets. Ensuring proper data provenance, open standards for metadata, and transparent algorithmic decision logs is essential for reproducibility and public trust. Robust frameworks for data sharing, privacy protection, and auditability must accompany ISL deployments.

\newpage
\bibliographystyle{plain}
\bibliography{main}

\newpage

\end{document}